\documentclass[pdflatex,sn-mathphys-ay,iicol]{sn-jnl}% Math and Physical Sciences Numbered Reference Style 
\usepackage{graphicx}%
\usepackage{multirow}%
\usepackage{amsmath,amssymb,amsfonts}%
\usepackage{amsthm}%
\usepackage{mathrsfs}%
\usepackage[title]{appendix}%
\usepackage{xcolor}%
\usepackage{textcomp}%
\usepackage{manyfoot}%
\usepackage{booktabs}%
\usepackage{multirow}
\usepackage{algorithm}%
\usepackage{algorithmicx}%
\usepackage{algpseudocode}%
\usepackage{listings}%
\usepackage{pifont}
\newcommand{\cmark}{\ding{51}}%
\newcommand{\xmark}{\ding{55}}%
\usepackage{float}
\usepackage{tcolorbox}
\tcbset{colback=blue!5!white, colframe=gray, fonttitle=\bfseries}
\usepackage{lipsum}
\usepackage{lineno}
\usepackage{todonotes}

\usepackage{makecell}
\usepackage{natbib}

\begin{document}

\title[\LARGE \bf
Grounding LLMs For Robot Task Planning  Using Closed-loop State Feedback]{\LARGE \bf
Grounding LLMs For Robot Task Planning   Using Closed-loop State Feedback}


\author*[]{\fnm{Vineet} \sur{Bhat}}\email{vrb9107@nyu.edu}
\equalcont{These authors contributed equally to this work.}
\author*[]{\fnm{Ali Umut} \sur{Kaypak}}\email{ak10531@nyu.edu}
\equalcont{These authors contributed equally to this work.}
\author[]{\fnm{Prashanth} \sur{Krishnamurthy}}\email{prashanth.krishnamurthy@nyu.edu}
\author[]{\fnm{Ramesh} \sur{Karri}}\email{rkarri@nyu.edu}
\author[]{\fnm{Farshad} \sur{Khorrami}}\email{khorrami@nyu.edu}

\affil[]{\orgdiv{Department of Electrical and Computer Engineering}, \orgname{NYU Tandon School of Engineering}, \orgaddress{ \city{Brooklyn}, \postcode{11201}, \state{NY}, \country{USA}}}

\abstract{Planning algorithms break complex problems into sequential steps for robots. Recent work employs Large Language Models (LLMs) to generate robot policies directly from natural language in simulation and real-world settings. Models such as GPT-4 generalize to unseen tasks but often hallucinate because they lack sufficient environmental grounding; supplying state feedback improves robustness. We introduce a task-planning method that uses two LLMs—one for high-level planning and one for low-level control—thereby raising task success rates and goal-condition recall. Our algorithm, BrainBody-LLM, is inspired by the human neural system, dividing planning hierarchically across the two LLMs and closing the loop with feedback that learns from simulator errors to fix execution failures. Implemented with GPT-4, BrainBody-LLM improves task-oriented success in the VirtualHome environment by 15$\%$ over competitive baselines. We also evaluate seven complex tasks in a realistic physics simulator and on a Franka Research 3 robotic arm, comparing our approach with other state-of-the-art LLM planners. Results show that recent LLMs can use raw simulator or controller errors to revise plans, yielding more reliable robotic task execution.}

\keywords{Robotic Task Planning, LLMs in Robotics, Closed loop feedback}



\maketitle

\section{Introduction}
\label{sec:intro}
LLMs, trained on corpora of internet-sourced text, have demonstrated capabilities akin to artificial general intelligence \citep{bubeck2023sparks}. The inherent world knowledge of LLMs, combined with their in-context learning ability is paving a new direction in robotic task planning. Prior work showed that LLMs can generate step-by-step instructions for complex tasks without any re-training or model parameter updates~\citep{huang2022language,pmlr-v205-huang23c, Sun2023AdaPlannerAP, 10160591, yao2023react,Ahn2022DoAI, progprompt, song2023llm,Kambhampati2024LLMsCP}. Despite promising results in diverse robotic tasks, grounding LLMs in a given environment is still an open problem. Consider the task - \textit{``Make me a coffee."} LLMs can decompose this problem into a sequence of steps-- `1. Walk to fridge,' `2. Grab milk,' and so on, through to `9. Serve cup of coffee.' Yet, these steps are not entirely executable in a real-world environment with physical constraints. Additional steps like `Switch on microwave' or `Open microwave doors,' are essential for task completion. Moreover, limitations like absence of milk or water should not hinder task execution. Robots must adapt to the environment,  refining plans towards successful task completion.

\begin{figure}[!t]
      \centering
      \includegraphics[width=\columnwidth]{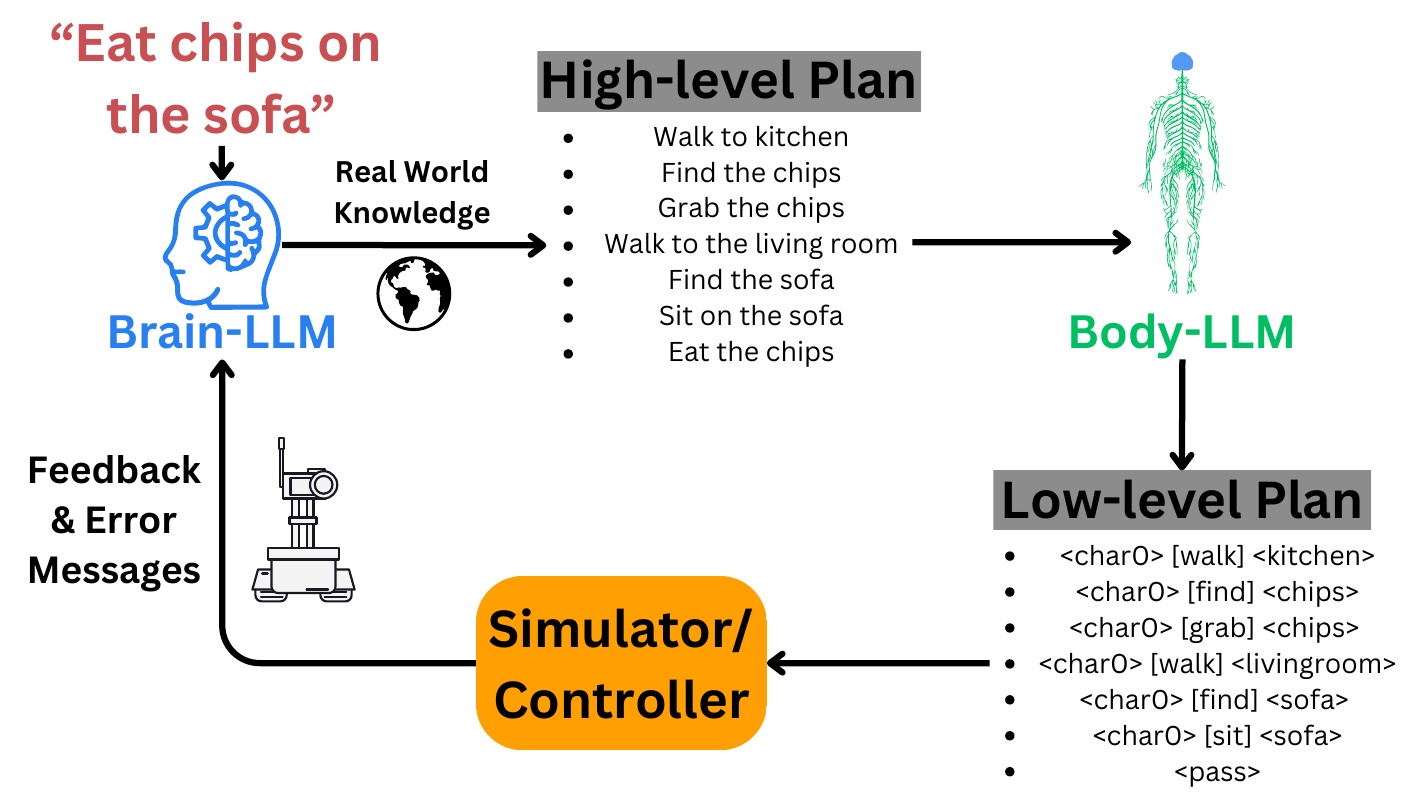}
      \caption{Illustration of how two LLMs work together in the proposed algorithm: The Brain-LLM splits the given task, `Eat chips on the sofa', into sequential steps using its real-world knowledge. The Body-LLM takes these steps one-by-one and determines executable actions. In instances where a corresponding action is not found in the environment, as demonstrated in the final step of this example, the Body-LLM outputs a $\textless$pass$\textgreater$ token.}
      \label{fig:GPT-4-plans}
\end{figure}

Thus, grounding LLMs in real-world scenes is essential. Incorporating environmental feedback into task planning allows error resolution in real-time, enhancing robot robustness and utility~\citep{pmlr-v205-huang23c}. In this paper, we introduce a novel planning algorithm that aims at mitigating two issues in existing methods: i) Our approach uses a simple prompting framework, with clear distinction of planning, feedback and execution components to avoid using expert defined heuristics and ii) Our method can integrate feedback from simulators, controllers or human intervention to guide an LLM planner for robust task execution, enhancing autonomy. Our contributions are:
\begin{enumerate}
\item A novel planning algorithm that uses a Two-LLM Agentic framework (Brain-LLM and Body-LLM) to derive executable actions from natural language instructions, leveraging a closed-loop state feedback mechanism for error resolution (Figure \ref{fig:GPT-4-plans}). 
\item Improving task-oriented success rate by 15\% average over existing state-of-the-art techniques in the VirtualHome Embodied Control environment (using a dataset of 80 tasks). BrainBody-LLM on average completes 81\% of all goal conditions for a given task.
\item Deployment and testing of our LLM based planner on the Franka Research 3 robotic arm, in 7 tasks of varied difficulties using a realistic physics simulator along with real robot experiments. 
\end{enumerate}

\section{Background and Related Work}
\label{sec:bg-related-work}

% \subsection{Foundational Models}

% Designing deep learning systems capable of comprehending and generating natural language has long been an attractive yet challenging task in the AI community. The emergence of transformers \citep{NIPS2017_3f5ee243} and the success of pre-trained language models such as BERT \citep{DBLP:journals/corr/abs-1810-04805} have paved the way for more advanced foundational models known as Large Language Models \citep{touvron2023llama, anil2023palm, openai2023gpt4}. These LLMs are trained on extensive textual data using masked language modeling and autoregressive text prediction. The discovery of in-context learning capabilities, a feature absent in earlier models, has laid the foundation for various applications of LLMs \citep{NEURIPS2020_1457c0d6}. In this study, we exploit these capabilities, particularly prompt engineering, which requires no gradient updates in the foundational model.

\subsection{Training LLMs for Robotic Policy Generation}

The real world knowledge contained within LLMs can be utilized in robots for common sense reasoning and generating language conditioned policies for task execution. Earlier works trained smaller LLMs like GPT-2 for agent task planning, treating it as a translation problem from natural language instructions to high level plans~\citep{jansen-2020-visually,micheli-fleuret-2021-language}. More recently, training LLMs in a multi-task setup for embodied control has demonstrated robust performance in planning, video captioning, video QnA and multi-turn dialogue~\citep{mu2024embodiedgpt}. Extremely large Vision Language Models (VLMs) trained on diverse robot demonstration data (comprising frame aligned RGB images and action vectors) can be used to directly convert vision-language input modalities to physical actions in the real world~\citep{rt12022arxiv,rt22023arxiv,Ahn2022DoAI,driess2023palme,kim2023lalm,belkhale2024rt}. Training VLMs is highly resource intensive, and accurate real world deployment requires customization through further environment specific training~\citep{Wake2023GPT4VisionFR}. Although methods like OpenVLA~\citep{kim24openvla} facilitate LoRA-based fine-tuning to reduce computational requirements, they cannot be deployed in zero-shot settings for novel environments without first collecting task-specific demonstrations and undergoing re-training. 

\subsection{Prompting LLMs for Task Planning}

Foundational models after GPT-3 have demonstrated excellent in-context learning abilities, which enables learning from examples in natural language without the need for expensive training and model updates. By choosing a suitable language prompt and some domain examples, LLMs can break down complex tasks into executable steps complying agent-environment constraints for high level planning~\citep{huang2022language,wang2023voyager,mahadevan_gen}. \cite{yao2023react} introduces an LLM framework for decision making in simulation environments.  LLMs can be used to learn from structured data such as Planning Domain Definition Language (PDDL) or Hierarchical Linear Temporal Logic (HLTL) and combined with heuristic search based planners~\citep{silver2022pddl,luo2023obtaining}. LLM generated PDDL world models can be used with domain independent planning models in robotic task planning~\citep{guan2023leveraging,Xie2023TranslatingNL}. Constructing PDDL world models is difficult in real world scenarios, where object and environment states are often revealed in runtime thus requiring human assistance for effective policy exploration. LLM generated plans are often in natural language and need to be converted to suitable action primitives for parsing and execution \citep{Xie2023TranslatingNL}. These action primitives must follow strict syntactic rules, as errors in these control statements can lead to downstream execution problems. A popular solution involves designing control functions in known programming languages like Python, which are then translated into rule-based action primitives~\citep{10160591, progprompt}. Our method uses simple action statements as execution primitives, but can be conditioned to output python code for robot execution, similar to previous studies in this area.

\subsection{Grounding LLMs to prevent Hallucination}

LLM generated plans sometimes consist of hallucinatory output, when the planner uses objects or actions that do not exist in the environment. This problem is mitigated by downstream filtering mechanisms that improve the correctness of LLM generated plans. \cite{knowno2023} uses a conformal prediction scheme to rank LLM generated plans for a current task. High level plans can be combined with motion planning, active preference learning and RL control policies for physical execution~\citep{wang2024apricot,dalal2024psl}. Visual perception greatly aids LLM generated sequence of skills through geometrically conditioned policy selection~\citep{Lin2023,shah2024bumble}. Combining LLMs with visual perception through VLMs  to construct 3D value maps can be used by high level planners for effective grounding in real world environments~\citep{huang2023voxposer}. Subsequent research has demonstrated application of such systems in navigation tasks through visual grounding~\citep{shah2022lmnav,zhou2023navgpt} or incrementally generated 3D scene graphs~\citep{rana2023sayplan,Rajvanshi_Sikka_Lin_Lee_Chiu_Velasquez_2024}. Decision making skills of goal driven LLM agents can be improved with runtime feedback~\citep{shinn2023reflexion} or affordance function based grounding~\citep{Ahn2022DoAI}. LLMs deployed in robots can incorporate feedback through natural language, and can use it to correct erroneous plans to enhance overall task success rate~\citep{10160591, Valmeekam2023OnTP}. Feedback from human intervention or simulator messages can improve real world alignment~\citep{guan2023leveraging}. Methods that use environmental feedback for task planning can be classified into two broad categories: static and dynamic planners. \textit{Static planners} use feedback to check whether the robot's environment satisfies the necessary conditions outlined in the generated plan, thus preventing execution errors. In this case, the feedback does not alter the generated plans but improves task oriented success rate~\citep{10160591, progprompt}. On the other hand, dynamic planners use feedback from the system to assert necessary conditions and to alter the plans,  improving their execution~\citep{pmlr-v205-huang23c, Sun2023AdaPlannerAP, yao2023react, song2023llm}. ProgPrompt, a well-known technique, was one of the pioneers of LLM-based planning for robotic tasks~\citep{progprompt}. Their algorithm used LLMs to break down complex high-level instructions into sequential steps by writing Python functions utilizing a known set of API skills. Assert statements (if-else) checked the existence of certain preconditions (e.g., whether the microwave door is open or the cup is on the table). This static planning mechanism improved both the success rate and task execution, as feedback from assert statements was utilized by LLMs to create robust plans. Our approach distinguishes itself from ProgPrompt and other conventional dynamic planners by employing a Two-LLM system designed to assimilate error messages from a simulator and environmental states in tandem for enhanced task planning. This allows our planner to understand errors in natural language and generate corrected plans for successful execution. Thus, we move one step further from ProgPrompt by replacing simple assertion-based precondition checks with natural language and real-time feedback. To the best of our knowledge, this work is the first to implement a dynamic planning algorithm for both the VirtualHome Simulator and the Franka Research 3 robotic arm.

\subsection{Closed loop LLM Planning}

Our work derives inspiration from previous methods in utilizing feedback by back-prompting LLMs for improved robotic task planning~\citep{Valmeekam2023OnTP,wang2023describe,skreta2023errors}. A hierarchical planner, which breaks down complex task instructions to intermediate planning and action primitives has been explored before~\citep{song2023llm}. However, we design our system to ensure minimal domain customization, and demonstrate applicability in both simulation and real world setups. \citet{rana2023sayplan} uses two distinct LLMs for 3D scene graph search and iterative re-planning using simulator feedback. Our approach does not rely on scene graphs, with planning and error resolution performed by the Brain-LLM and action mapping performed by Body-LLM. \citet{pmlr-v205-huang23c} implements a closed loop feedback integrating visual perception and language conditioned robotic skills. In their approach, the LLM planner has access to vision models providing information about the environment, and it further disambiguate planning confusion by asking questions to a human operator. Our contributions lie in an effective and autonomous planning approach that eliminates any human-in-the-loop scenario by reinforcing feedback through error messages. There are a number of works that integrate visual grounding in planning with VLMs for object manipulation~\citep{Wu2023TidyBotPR,Liu2024OKRobotWR,huang2023voxposer,chen2023llm,liu2023reflect} and navigation tasks~\citep{shah2022lmnav,zhou2023navgpt}, and we leave such expansion of our planning methodology for future work. Our work can also be integrated with skill learning techniques through human-guided imitation learning as demonstrated in \cite{parakh2023humanassisted}. 

A two-LLM system for generating symbolic and geometric relationships for task and motion planning has shown effectiveness in robotic environments~\citep{10342169}. Our method distinguishes itself by avoiding any human expert intervention or customization for filtering, allowing environmental feedback and error messages to automatically guide the planning LLM towards an accurate and correct real world action. Traditional simulators---human-crafted world models---have long supplied the environmental feedback and state information used for robot planning. Recent advances in camera-based perception, however, offer far richer visual context for LLM-driven planners~\citep{molmo2024,ravi2025sam}. Since our algorithm is perception-agnostic, it incorporates these improvements without modification, providing a straightforward yet highly scalable framework for end-to-end, language-guided robot control. \citet{Silver_Dan_Srinivas_Tenenbaum_Kaelbling_Katz_2024} uses two LLMs for planning and correction using execution feedback, but relies on PDDL models for effective grounding, whereas our approach eliminates this dependency and can be used in real world experiments where objects and environment states are not known during task planning. The closest to our work is \cite{li2023interactive}, which also uses an LLM for high level step by step task planning, followed by another LLM for converting plans to low level functional APIs for robot control. Their method is capable of re-planning based on human-in-the-loop feedback, while ours uses task errors (which can be through simulator or a human feedback) to re-plan.

\begin{algorithm}
\caption{BrainBody-LLM}
\begin{algorithmic}[1]
\small
\State \textbf{Input:}
 \State $T$ \Comment{Task description}
\State $K$ \Comment{Max feedback loops}
\State \textbf{Functions:}
\State $ \phi(\text{task}, \text{feedback})$ \Comment{Brain-LLM}
\State $\theta(\text{high level plan step})$ \Comment{Body-LLM}
\State $\pi(\text{low level action})$ \Comment{Actuator}
\State $\rho(\text{action result})$ \Comment{Simulator/Perception}
\State \textbf{Start}
\State $k \gets 0$ \Comment{feedback loop counter}
\State $f \gets $none \Comment{no feedback in the beginning}
\State $hlp \gets \phi(T, f)$ \Comment{high Level Plan}
\State $i \gets 0$
\While{$i \neq length(hlp)$}
    \State $step \gets hlp[i]$ \Comment{ith step of high level plan}
    \State $\delta \gets \theta(step)$ \Comment{low level Action}
    \If{$\delta \neq <\text{pass}>$}
        \State $action\_result \gets \pi(\delta)$
        \If{$action\_result \neq \text{success}$}
            \If{$k \neq K$} \Comment{Update plan}
            \State $f \gets \rho(action\_result)$ \Comment{feedback}
                \State $hlp \gets \phi(T, f)$
                \State $k \gets k + 1$
            \Else
            \State $i \gets i + 1$ \Comment{go to next step}
            \EndIf
        \Else \Comment{action is successful}
        \State $i \gets i + 1$ \Comment{go to next step}
        \EndIf
    \Else \Comment{no primitive for the current action}
    \State $i \gets i + 1$ \Comment{go to next step}
    \EndIf
\EndWhile
\end{algorithmic}
\label{algo:bbllm}
\end{algorithm}

\section{Our Proposal: BrainBody-LLM}
\label{sec:algo2}

The human brain is a sophisticated information processing system comprising a network of billions of neurons. Artificial neural networks aim to mimic brain functions, but a significant difference lies in the information processing method: the human brain functions continuously, in contrast to the discrete operation of computer algorithms \citep{korteling_et_al,aimone_et_al}. To emulate critical cognitive functions of the human brain, particularly in terms of feedback mechanisms and adaptability, we developed the BrainBody-LLM algorithm. 

Our algorithm consists of two LLMs, each with a distinct contribution to the overall task execution pipeline. The Brain-LLM is designed to decompose a given task into high-level execution plans in natural language. High-level plans are converted by the Body-LLM into low level robot control commands, with a pre-defined syntax for environmental execution. During runtime, any error messages provided by the simulation environment, motion planning controller or human feedback can be relayed to the Brain-LLM, which then generates an updated plan from the current step to complete the original task. The updated plan is subsequently converted to control statements by the Body-LLM. Our method uses iterative planning and feedback to create a closed loop pipeline. Our complete framework of integrating two LLMs is described in Algorithm \ref{algo:bbllm}. The framework begins by passing the task description \textbf{$T$} and initial empty feedback \textbf{$f$} to the Brain-LLM \textbf{$\phi$}, producing a high-level plan \textbf{$hlp$}.
The loop index \textbf{$i$} iterates through each step of this plan, and the Body-LLM \textbf{$\theta$} converts the current step into a low-level action \textbf{$\delta$}, which the actuator policy \textbf{$\pi$} executes to yield an \textbf{$action\_result$}.
If the result is not a success, the perception/simulator module \textbf{$\rho$} extracts new feedback \textbf{$f$}; provided the feedback-loop counter \textbf{$k$} is below the maximum \textbf{$K$}, this feedback is sent back to \textbf{$\phi$} for on-the-fly re-planning, otherwise the algorithm proceeds to the next step.
The procedure continues until every plan step succeeds or no further primitives are available, giving a simple yet flexible two-LLM framework that couples high-level reasoning with low-level control while supporting up to \textbf{$K$} adaptive replanning cycles.
All LLM prompts in our approach include in-context learning examples that inform the planner about the environmental constraints (e.g., the microwave is closed or the TV is on) and provide accurate, manually curated task-plan example pairs. BrainBody-LLM uses these examples to uncover patterns and understand object-action relationships to generate plans for unseen tasks. Moreover, feedback through error messages further grounds the planner in the real world by informing it about erroneous steps while explaining reasons for the failure.

\noindent \textbf{State Representation:} Our algorithm assumes access to a continuous state representation of the environment during task execution. State information is utilized to recover from execution failures, enabling the robot to realign and complete the task. We represent the scene at time~$t$ by the state set
\[
S_t = \bigl( \mathcal{O}_t,\, \mathcal{A}_t,\, \mathcal{R}_t \bigr),
\]
where $\mathcal{O}_t = \{o_1,\dots,o_{|\mathcal{O}_t|}\}$ denotes the objects currently present,  
$\mathcal{A}_t = \{a_1,\dots,a_{|\mathcal{A}_t|}\}$ the robot actions executable in this state, and  
$\mathcal{R}_t = \{(s{:}p{:}o)_k\}_{k=1}^{|\mathcal{R}_t|}$ the relational triplets describing object properties and pairwise relations—for example, $(\textit{fridge}:{\textit{isOpen}}:{\textit{true}})$ or $(\textit{cup}:{\textit{on}}:{\textit{table}})$. In simulation environments, such states are typically available through built-in perception modules. In real-world trials, these states can be constructed using visual perception techniques. In this work, the terms `environmental information' and `state representation' are used interchangeably. For instance, \cite{guan2024task} propose leveraging vision-language models (VLMs) to identify undesirable robot behaviors from demonstrations, providing feedback to refine planning. However, accumulating such states in real-time can create bottlenecks in the deployment of large language model (LLM) algorithms. Recent advancements in real-time visual tracking algorithms, such as those by~\cite{yang2024samurai}, offer potential solutions to mitigate these issues. Our contribution lies in developing an algorithm utilizing these states to enhance planning, with the accumulation of state information being outside the scope of this work.

We design three prompts to ground the LLMs within the environment and produce output that is compatible with real-world or simulation experiments.  Each prompt addresses a key component of our pipeline: Planning, Execution and Feedback. 

\subsection{Planning}

\begin{figure*}[!h]
  \centering
  \small
  \begin{tcolorbox}[title=Planning Prompt for Brain-LLM]
    You are in the command of a mechanical robot. Your task is to split a given task into high-level steps that can be executed by the agent in the current environment. Each output step should be executable by the agent using available actions.\\
    
    Some examples of Task Instruction - Step-by-step plan pairs are given below: \\
    
    \{in\_context\_learning\_examples\}\\
    
    You have the following objects in scene: \{object\_list\}.
    The list of available actions are - \{actions\_available\}. \\
    
    Use the information above to create the step-by-step plan for the given task instruction. Remember to only use the above objects and the available actions. Do not combine intermediate steps to generate compound steps. Make sure to complete all the steps needed to finish the task. Your reply should always start with ``0:"\\ 
    
    \textbf{Environment Information:} \{environment\_information\}\\
    \textbf{High-level Instruction:} \{task\_input\}\\
    \textbf{Step-by-step Instructions:}
  \end{tcolorbox}
  
  \caption{Format of the planning prompt used in our experiments. The planning prompt tunes LLM outputs to meet environmental constraints while generating step-by-step task execution plans. In-context learning examples of high-level tasks and their corresponding subtasks, along with a list of available objects and actions, are needed. This enables the LLM to learn patterns from the examples and create plans for unseen tasks based on the robot's current environment.}
  \label{fig:planning-prompt}
\end{figure*}

The planning prompt is designed to provide Brain-LLM with all necessary information to understand the robot's environment, the available actions within the simulator or real-world setup, and some examples to facilitate learning the planning task in a few-shot setting. These examples can be manually crafted or selected from an appropriate dataset. Each example consists of a tuple: \textit{(Environment Information, Input Task, High Level Plans)}. Given that LLMs are trained on extensive real-world datasets, they rapidly learn patterns between tasks and their high-level execution plans, as well as the grounding relations between objects and actions required for executing a given task. Using common sense reasoning, they also learn to hierarchically generate plans for a task that are consistent with previously generated plans for similar tasks while ensuring temporal continuity. Figure \ref{fig:planning-prompt} illustrates the format of our prompt, shortened for brevity.

\begin{figure*}[h]
  \centering
  \small
  \begin{tcolorbox}[title=Execution Prompt for Body-LLM]
        You are in control of a robot. Your task is to create a single line of program in a described format based on the instructions provided to you. 
    Agents interact in environments via programs which are instructions that describe which actions each agent should do, and with which objects. Each line of program has the following format -\\ 
    
    Action Plan: \{command\_syntax\} \\

    Some examples of plan - action program pairs are given below - \\

    \{incontext\_examples\_execution\}\\

    You have the following objects in scene: \{object\_list\}. The list of available actions are - \{actions\_available\}.

Return a suitable action program for the provided plan. Remember to only use the actions in the available action list and use objects in the provided object list. If you are not sure what the output should be, it is always better to $\textless$pass$\textgreater$ instead of creating a wrong action. If you do not find an available action or an object for the given sub-task, you should simply output $\textless$pass$\textgreater$.\\

\textbf{Description}:  \{input\} \\
\textbf{Action Plan}:
  \end{tcolorbox}
  \caption{Format of the execution prompt used in our experiments. The execution prompt tunes the Body-LLM to generate appropriate control statements in the required syntax for a given plan created by the Brain-LLM.}
  \label{fig:execution-prompt}
\end{figure*}

\subsection{Execution}

The Body-LLM is responsible for sequentially generating low level executable policies based on the natural language plans created by the Brain-LLM. The execution prompt includes examples of natural language steps paired with robot control statements, helping Body-LLM learn the associations necessary for task execution. Some examples of the generated low-level policies are shown in Table \ref{tab:bodyllm-examples}. The prompt also introduces a unique token, $\textless$pass$\textgreater$, which Body-LLM uses when a natural language plan lacks a realizable action in the environment. In our pipeline, the $\textless$pass$\textgreater$ token allows skipping steps, which is crucial for preventing execution errors and avoiding oscillatory behavior caused by unavailable execution statements. This helps prevent erroneous plans from repeating due to the limited context window of LLMs. Figure \ref{fig:execution-prompt} illustrates our execution prompt.

\begin{figure*}[h]
  \centering
  \small
  \begin{tcolorbox}[title=Feedback Prompt for Brain-LLM]
    You are in the command of a robot. Given a high level task and associated subtasks, a controller executed the commands by converting them to robotic syntax for object manipulation. However, not all subtasks were successful, and your job is to examine an error in execution, and suggest a revised action plan in continuation with previously executed commands. \\

    Some examples of high level tasks, generated subtasks, error step and error message are given below. Go through them to understand the type of errors that are encountered, and learn how the revised plan can solve the encountered error -\\ 

    \{incontext\_examples\_error\_resolution\} \\
    
    For the given task: \{input\}, a generated Initial Plan was: \{init\_plan\}. The robot received the following feedback message: \{feedback\_message\} \\

    You have the following objects in scene: \{object\_list\}. The list of available actions are - \{actions\_available\}. \\

    Use the information above to create an updated step-by-step plan for the given task such that this error does not occur again. Remember to only use the above objects and the available actions. Do not combine intermediate steps to generate compound steps. Your response should always start with numbering from the error step:\\
    
    \textbf{Environment Information:} \{env\_information\}\\
    \textbf{High-level Instruction:} \{input\}\\
    \textbf{Explanation:}\\
    \textbf{New plan:}
  \end{tcolorbox}
  \caption{Format of the feedback prompt used in our experiments. The feedback prompt informs the LLM of an execution error and provides examples of how similiar errors can be resolved. The LLM learns from these examples, and uses the given environmental conditions, available actions and creates a new updated plan to resolve the error, conditioned on already executed plans before the error step.}
  \label{fig:feedback-prompt}
\end{figure*}

\subsection{Feedback}

Error resolution in high level task planning has been explored through three primary sources of feedback: rule-based heuristics~\citep{huang2022language,10342169}, simulator feedback~\citep{rana2023sayplan,Sun2023AdaPlannerAP,Silver_Dan_Srinivas_Tenenbaum_Kaelbling_Katz_2024}, and human feedback~\citep{parakh2023humanassisted,pmlr-v205-huang23c}. Our method can integrate feedback from any of these sources to prompt the Brain-LLM to update its plans. The feedback prompt informs the Brain-LLM about the occurrence of an execution error and any associated error messages from the controller, simulator, or a human. Examples of error instances and corresponding solutions through new plans are also provided. Similar to the planning prompt, these in-context examples can be chosen either manually or from a suitable dataset. Each example consists of the tuple \textit{(Error Message, Explanation, Updated Plans)}. \textbf{In our experiments, error messages are collected from the simulator or controller and indicate which action was not executed and the reason.} Often, the raw error message is unreadable, so the \textit{Explanation} explains the problem to the Brain-LLM in natural language and why a particular step failed. LLMs have been shown to improve performance with such chain-of-thought reasoning steps~\citep{NEURIPS2022_9d560961}. Finally, the updated plans are constrained to start from the error step and are conditioned on plans completed before the error step. In real-world applications, it is not feasible to restart the entire task since the robot might have already executed some plans. Thus, we constrain the Brain-LLM to always generate updated plans from the current error step. When deployed to resolve a particular error, the feedback prompt encourages the LLM to first generate a natural language reasoning for the failure, followed by step-by-step plans for error resolution. Figure \ref{fig:feedback-prompt} shows the prompt we use for Brain-LLM. In this work, we do not incorporate visual perception models. However, our framework is designed to be easily extendable to include such models, such as Grounding DINO~\citep{liu2023grounding}, as visual perception modules to inform the BrainLLM about the environment. Integrating these models would enable continual feedback for reversible tasks. For instance, if a task involves pouring water into a glass but the water is accidentally spilled on the table, continual visual perception could provide feedback to the BrainLLM for re-planning. To simplify our approach, we assume that the robot has prior knowledge of object locations. This assumption is practical and can be relaxed using modern visual perception techniques.

\section{Experiments with VirtualHome}
\label{sec:experiments}

To evaluate our BrainBody-LLM approach, we first utilize a simulator and a perception module that execute commands generated by our LLMs and return environment states and error messages. We use a widely-adopted robotic task simulation software, ensuring the details provided allow for replication of our results.

\subsection{VirtualHome: Simulator For Embodied Control}

VirtualHome (VH) serves as our robotic control software, simulating a Human-In-A-Household scenario with support for multiple agents~\citep{puig_virtualhome}. The simulator features a variety of interactive household objects with predefined states like ``open," ``closed," ``on," and ``off." VH represents the agent as a humanoid avatar capable of interacting with these objects via low level control statements. Additionally, the simulator includes an in-built perception module that provides real-time information about objects in the scene, their states, and positions. Success and failure messages (with reasoning) for robotic commands are also provided. We use the environmental states and error messages from the in-built simulator as input to the feedback prompt described in Figure \ref{fig:feedback-prompt}. For our experiments, we use VirtualHome v2.3.0.

\subsection{Dataset}

We observed that many tasks from the original VH dataset, which were manually annotated, are no longer executable in the latest simulator version. Therefore, we utilize the train-validation-test subset of samples as used in \citet{progprompt}. Each sample comprises a tuple: a task description, a step-by-step plan (both in natural language), and corresponding VH commands. The dataset is divided into training, validation, and test splits, containing 35, 25, and 10 tuples, respectively. Our planning prompt uses examples of task descriptions and high-level plans from the dataset, whereas the execution prompt uses high-level plans and their corresponding VH commands as in-context learning examples.

\subsection{Models Used}

Since our approach relies on pre-trained foundational models that embody real-world knowledge and support incontext learning through prompting, we use three popular LLMs trained on vast amounts of data: PaLM 2 text-bison-001 \citep{anil2023palm}, GPT-3.5 \citep{NEURIPS2020_1457c0d6}, and GPT-4 \citep{openai2023gpt4}. These models are accessed via their API calls.  Furthermore, since Progprompt achieves optimal performance with the DAVINCI-003 model \citep{NEURIPS2020_1457c0d6}, a variant of GPT-3, we also report its performance using the DAVINCI-003 backbone.

\subsection{Baselines}

For fair comparisons, we selected baselines tested in similar scenarios—without visual perception-based grounding or human feedback. Our study benchmarks against two recent works that use LLMs for agent control in the VH environment~\citep{progprompt,huang2022language}. Additionally, we developed a baseline model (Baseline-LLM) to translate task descriptions, expressed in natural language, directly into VH commands for execution without any intermediate high-level plans. Unlike the BrainBody-LLM, the Baseline-LLM uses a single LLM without any environmental feedback but with the same training dataset. These comparisons will highlight the efficacy of our BrainBody-LLM in the context of LLM-based embodied control. Some in-context learning examples used in our prompt structure are described in Appendix \ref{app:prompts-vh}. 

\subsection{Evaluation Metrics}
We use three metrics for evaluating the plans produced by BrainBody-LLM and the baselines as used in previous works~\citep{huang2022language,progprompt}.

\noindent\textbf{Executability (EXEC)} measures the proportion of steps in the plan that are executable within the VH environment. This metric does not assess the correctness of individual action plans in achieving the final goal. Thus, while EXEC primarily evaluates the performance of the Body-LLM in our approach, Brain-LLM generated plans influence this score too, as impractical plans result in misaligned commands.

\noindent \textbf{Goal Conditions Recall (GCR)} measures the percentage of satisfied goal conditions for a given task. It is calculated as the ratio of satisfied goal conditions to the total required goal conditions. For example, consider the task of bringing a coffee pot and a cupcake to the coffee table in VirtualHome. Both the coffee pot and the cupcake must be on the coffee table at the end of the task execution. If neither item is carried to the table, the GCR is 0. If one item is moved to the table, the GCR is 0.5. If both items are successfully moved to the table, the task is considered successful, and the GCR is 1.

\noindent \textbf{Success Rate (SR)} quantifies the rate of successful tasks completed by an algorithm. For a task to be counted as successful, all task-relevant goal conditions should be satisfied. In other words, the SR of a task is 1 if and only if its GCR is 1. To calculate the overall SR, an average is taken over the SR of the tasks in a dataset.

% \noindent \textbf{Addressing a Key Limitation of Evaluation Metrics:} The SR metric above considers relations amongst all objects in the scene, rendering it a stringent measure. However, true success rate should only account for changes in objects pertinent to the task. For instance, in the task ``Make a Sandwich," if the robot alters the state of the light, and such an action is not part of the ground truth plan, the plan should not be penalized.  Thus, we introduce \textit{Corrected Success Rate (CSR)}, considering only relations and states from required objects for the task. We annotate these objects for each task in the test set. CSR more accurately reflects effectiveness and is a better indication of success rate.

% is the difference between the desired final state and the state achieved after executing all generated commands. This metric uses two variables: Unsatisfied Goal Conditions (UGC), which tracks the disparities between the predicted and actual final states, and Total Goal Conditions (TGC), which records the differences in object states and relations from initial to final state: %The GCR is formulated as follows:
% \begin{equation}
% GCR = 1 - \frac{UGC}{TGC}.
% \label{eq:GCR}
% \end{equation}

\subsection{Observations and Results}
% \label{sec:obs-results}

\begin{table*}[!t]
\caption{Comparing LLM-based planning algorithms for the Virtual Home Simulator Environment. We do not use the DAVINCI-003 LLM since it has been deprecated.}
\label{tab:baseline-results}
\centering
\begin{tabular}{@{}lccccc@{}}
\toprule[1pt]
\textbf{Technique} & \textbf{Feedback} & \textbf{LLM} & \textbf{SR} & \textbf{GCR} \\
\midrule[1pt]
 \citet{huang2022language} & \xmark & GPT-3 & 0.00$\pm$0.00 & 0.21$\pm$0.03\\ 
\midrule[0.5pt]
\multirow{3}{*}{ProgPrompt\citep{progprompt}} & \multirow{3}{*}{\cmark} & GPT-3 & 0.34$\pm$0.08 & 0.65$\pm$0.05\\ & & DAVINCI-003 & 0.47$\pm0.15$ & 0.74$\pm$0.07\\ & & GPT-4 & 0.37$\pm0.06$ & 0.64$\pm0.02$ \\  
\midrule[0.5pt]

\multirow{2}{*}{Base-LLM} & \multirow{2}{*}{\xmark} & PALM & 0.30$\pm$0.00 & 0.60$\pm$0.07\\
&  & GPT-3.5 & 0.16$\pm$0.05 & 0.59$\pm$0.01\\  \midrule[0.5pt]
\multirow{3}{*}{BB-LLM} &\multirow{3}{*} {\xmark} & PALM & 0.30$\pm$0.08 & 0.54$\pm$0.13 \\ 
& & GPT-3.5 & 0.26$\pm$0.15 & 0.69$\pm$0.09 \\ 
& & GPT-4 & 0.38$\pm$0.13 & 0.73$\pm$0.07 \\ \midrule[0.5pt]
\multirow{3}{*}{BB-LLM} & \multirow{3}{*}{\cmark} & PALM & 0.40$\pm$0.00 & 0.55$\pm$0.01\\
& & GPT-3.5 & 0.36$\pm$0.18 & 0.70$\pm$0.05 \\
& & GPT-4 & \textbf{0.54$\pm$0.09} & \textbf{0.81$\pm$0.04}\\
\bottomrule[1pt]
\end{tabular}
\end{table*}

Our test set consists of 10 tasks in Virtual Home as described in Table \ref{tab:baseline-results}. For each experiment, we run our algorithm five times with a temperature of 0.5 and report average scores across the runs. We study three variants of our algorithm: BrainBody-LLM (with feedback), BrainBody-LLM (without feedback) and Base-LLM (without Body-LLM and intermediate high level planning). 

Obtained results demonstrate that feedback based planning and correction methods such as ProgPrompt and BrainBody-LLM (with feedback) outperform baselines that do not use feedback. Without feedback based error resolution, non-realizable plans are skipped which leads to unsatisfied goal conditions. ProgPrompt uses assert statements (if-else conditionals) to validate necessary environmental conditions before executing a given step. BrainBody-LLM (with feedback) uses error messages (in natural language) to guide the Brain-LLM towards a better plan and subsequent improved goal conditions satisfaction. 
Our approach exhibited a significant enhancement in the SR metric, outperforming ProgPrompt with its highest-scoring LLM backend, Davinci-003, while achieving improved GCR when compared with the same LLM backend. Modifying our backend LLM from GPT-3.5 to GPT-4 improves our scores, motivating further improvements as LLMs become more powerful and their context window increases. In our approach, GPT-4 was the top performer, followed by PaLM 2 text-bison-001 and GPT-3.5. We also observe that BrainBodyLLM outperforms the Base-LLM baseline significantly on both SR and GCR when using the same LLM backend. Base-LLM was prompted to directly output low level control policies from task instructions. This shows that using two LLM calls for high level planning and low level policy generation leads to better results since LLMs are better at comprehending natural language prompts rather than functional APIs in their in-context learning examples. This motivates us to utilize a hierarchical structure for task planning, separating high-level task planning from low-level policy generation.

\begin{figure}[!h]
      \centering
      \includegraphics[width=\columnwidth]{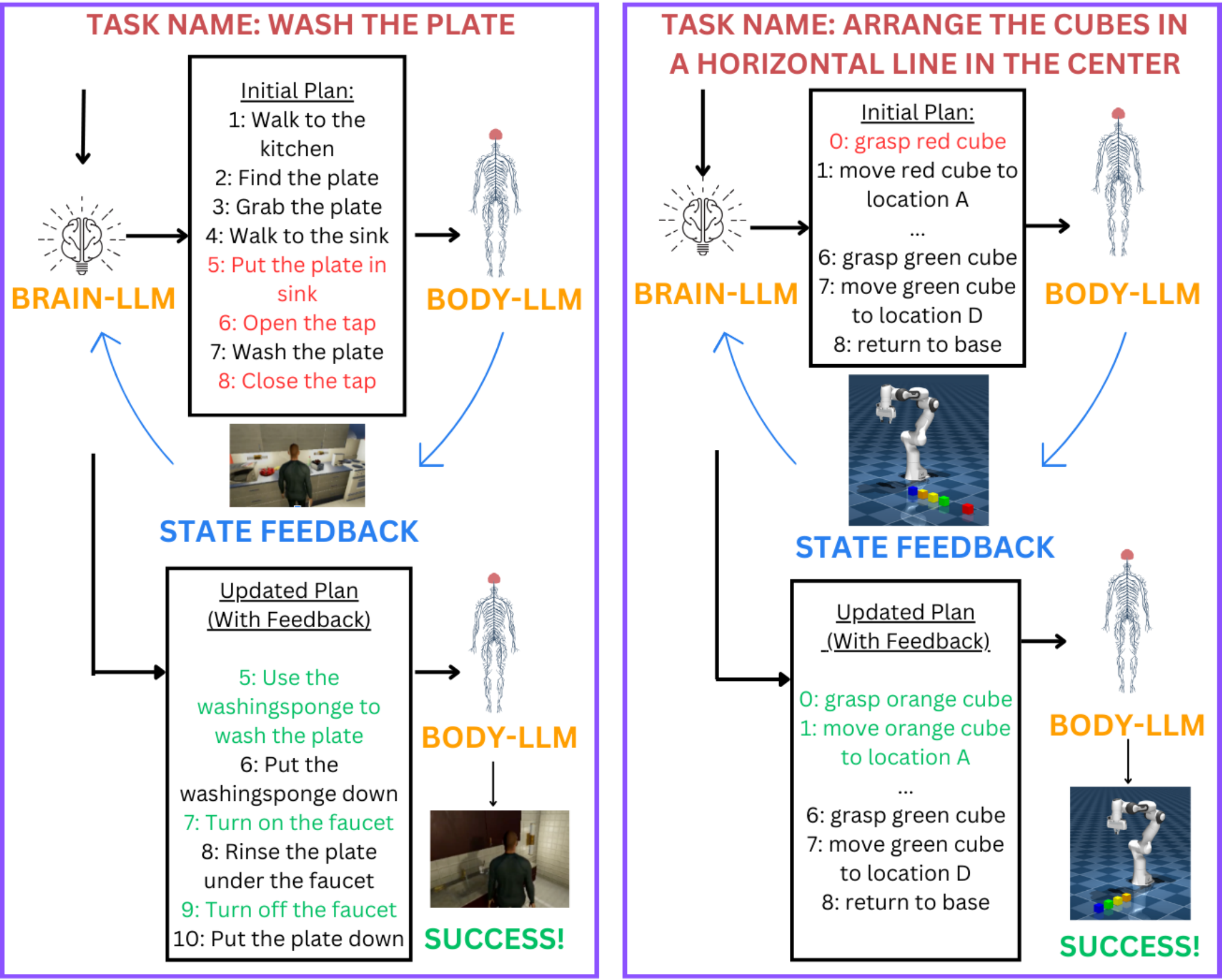}
      \caption{GPT-4 planning with and without feedback: Example 1 and Example 2 shows refinement of plans with context cues in VH and Franka Arm Simulation, respectively.}
      \label{fig:GPT-4-plans-feedback}
\end{figure}

\begin{table*}[!ht]
\caption{Best Score Comparison across models, showcasing GPT-4's superior execution with feedback, achieving perfect scores in 70\% of tasks, and highlighting its innovative problem-solving.}
\label{tab:technique_comparison}
\centering
\scalebox{0.85}{
\begin{tabular}{@{}l|ccc|ccc|ccc@{}}
\toprule[1pt]
\multirow{2}{*}{\textbf{Task}} & \multicolumn{3}{c|}{\textbf{PaLM 2 text-bison-001}} & \multicolumn{3}{c|}{\textbf{GPT-3.5}} & \multicolumn{3}{c}{\textbf{GPT-4}} \\
\cmidrule(l){2-10} 
 & SR & GCR & EXEC & SR & GCR & EXEC & SR & GCR & EXEC \\
\midrule[1pt]
Bring coffeepot and cupcake to coffee table & 0 & 0.08 & 0.36 & 0 & 0.31 & 1 & \textbf{1} & 1 & 0.50\\
Brush teeth & \textbf{1} & 1 & 1 & 0 & 0.28 & 0.43 & \textbf{1} & 1 & 0.50 \\
Eat chips on the sofa & 0 & 0.08 & 0.75 & 0 & 0.96 & 0.88 & \textbf{1} & 1 & 0.75 \\
Make toast & 0 & 0.98 & 0.38 & \textbf{1} & 1 & 0.57 & \textbf{1} & 1 & 0.42 \\
Microwave Salmon & \textbf{1} & 1 & 0.90 & 0 & 0.84 & 0.69 & 0 & 0.90 & 0.89 \\
Put salmon in the fridge & \textbf{1} & 1 & 1 & \textbf{1} & 1 & 0.88 & \textbf{1} & 1 & 0.88 \\
Throw away apple & 0 & 0.16 & 0.75 & \textbf{1} & 1 & 0.82 & 0 & 0.16 & 0.57 \\
Turn off light & \textbf{1} & 1 & 1 & \textbf{1} & 1 & 1 & \textbf{1} & 1 & 0.75 \\
Wash the plate & 0 & 0.16 & 1 & 0 & 0.13 & 0.54 & 0 & 0.10 & 0.91 \\
Watch TV & 0 & 0 & 0.67 & \textbf{1} & 1 & 0.80 & \textbf{1} & 1 & 0.80 \\
\bottomrule[1pt]
\end{tabular}
}
\end{table*}

BrainBody-LLM works best with the GPT-4 backend, utilizing simulator feedback for error resolution demonstrating a perfect score in 7-out-of-10 tasks (Table \ref{tab:technique_comparison}). Improved task-object mapping (plate $\rightarrow$ plate and not dish bowl, bread $\rightarrow$ breadslice) and spotting non-executable commands (using $\textless$pass$\textgreater$ token) boosts EXEC. GPT-4 and feedback-enhanced planning reveals creative problem-solving. Figure \ref{fig:GPT-4-plans-feedback} (LHS) shows how feedback reinforces task planning in VH environment. By utilizing error messages from the simulator, GPT-4 can generate revised plans that avoid repeating previous mistakes. The extensive world knowledge embedded in these models empowers them to not only comprehend the problem but also to formulate innovative and unexpected solutions, surpassing human intuition in certain cases. 

LLMs are prone to hallucinations, and using two LLMs can cause compounded errors. The $\textless$pass$\textgreater$ token allows the BodyLLM to skip a hallucinated or infeasible high-level plan created by the BrainLLM. However, we observed that in many cases, the BodyLLM attempts to devise a suitable action plan, which is often non-executable. Compared to ProgPrompt, which uses assertion-based pre-condition checks before executing a command, our method scores lower in EXEC. Incorporating assertion-based checks into our framework will help further improve EXEC. Notably, human-level EXEC, calculated by manually assigning action statements to high-level instructions, stands at 94\%. This highlights the challenges in utilizing simulation environments like VirtualHome for our experiments, as many logical and syntactically correct statements were not executed due to simulator limitations in object properties.
% We encountered issues with the simulation software related to the design and properties of certain objects. Some control statements, although satisfying the logic and syntax defined by the simulator, were not executed due to missing backend software capabilities, leading to the repetition of previous steps and oscillation of the plan. With realistic object properties, our model could likely succeed in the remaining tasks as well. 

Our closed-loop feedback mechanism for LLM-based planning is potentially susceptible to oscillatory behaviors that occur when a Body-LLM-generated action primitive is syntactically correct but not executable due to physical constraints in the environment. An unresolved plan repeats itself and compounds the error through multiple feedback loops. Increasing the number of feedback loops does not always enhance the model's performance. Incorporating additional sensor modalities into the LLM-planning framework can help prevent such issues. For instance, an image-level scene graph description of the environment can inform the LLM of existing conditions, allowing it to revise erroneous plans accordingly.

\begin{figure}[!h]
      \centering
      \includegraphics[width=\columnwidth]{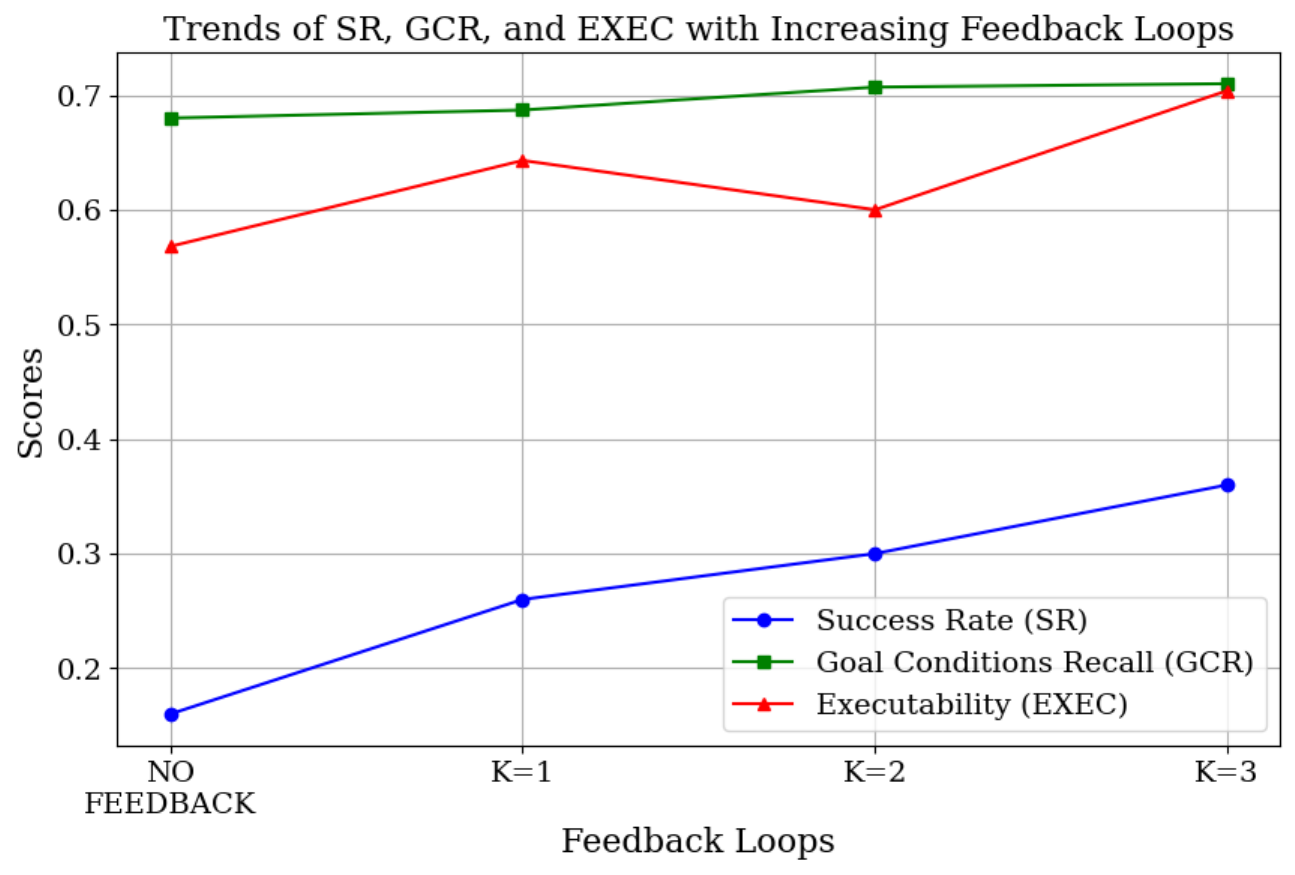}
      \caption{Impact of Feedback Loops on Evaluation Metrics. The graph demonstrates a clear upward trend in all evaluation metrics as the number of feedback loops increases, without significant fluctuations. Notably, upon incrementing the value of K beyond 3, the metrics stabilize around a consistent value. This value aligns with the results reported at K=3; therefore, data for values higher than K=3 are not included in the analysis. While the graph shows results with GPT 3.5, similiar trend is observed with GPT 4.}
      \label{fig:ablation-fig}
\end{figure}

\noindent \textbf{Ablation Study on Feedback Loops:} The two LLMs in our approach interact with each other through feedback messages from the simulator. Without feedback, error messages from incorrect plans are not utilized for correction. We ablate the number of feedback loops \(K\) (Algorithm \ref{algo:bbllm}) used in our experiments. Our results indicate a consistent improvement across all evaluation metrics as the number of feedback loops increases, without significant fluctuations, as depicted in Figure \ref{fig:ablation-fig}. 

% & Actions like put and putin require two target objects - primary and secondary. A BERT based matching method cannot correctly identify the primary and secondary objects, and uses the wrong action whereas BodyLLM uses in-context examples and advanced reasoning abilities to decipher the command.

\begin{table*}[!h]
\caption{Comparisons highlighting robustness of BodyLLM generated action statements. Actions like \textit{``put"} and \textit{``putin"} require detecting primary and secondary object, which the BodyLLM excels at due to better language reasoning capabilities. The second example shows how a simple high level plan can generate ambiguity in Embedding matching output, leading to failed execution. Third example highlights how BodyLLM correctly skips plans that are not executable, whereas embedding matching creates an erroneous action plan. In all examples, output generated by BodyLLM was executable and correct.}
\small
\label{tab:bodyllm-examples}
\centering
\begin{tabular}{@{}ccc@{}}
\toprule[1pt]
\textbf{High level Plan} & \textbf{BodyLLM} & \textbf{Embedding Matching}\\
\midrule[0.5pt]
\makecell{Put toothpaste on \\ the toothbrush} & $\textless$putin$\textgreater$ $\textless$toothpaste$\textgreater$ $\textless$toothbrush$\textgreater$ &  $\textless$put$\textgreater$ $\textless$toothbrush$\textgreater$ $\textless$toothpaste$\textgreater$ \\ 
\midrule[0.5pt]
Close the fridge &  $\textless$close$\textgreater$ $\textless$fridge$\textgreater$ &    $\textless$switchoff$\textgreater$ $\textless$fridge$\textgreater$ \\ 
\midrule[0.5pt]
Wait for the toast to be ready &  $\textless$pass$\textgreater$ &  $\textless$drink$\textgreater$ $\textless$toaster$\textgreater$ \\ 

\bottomrule[1pt]
\end{tabular}
\end{table*}

\noindent \textbf{Need for Body-LLM:} In our approach, the Body-LLM performs an association task between natural language high-level plans and low-level control statements for task execution. Such association tasks can also be performed with a simple text-matching algorithm that aligns natural language steps with a fixed set of executable actions and objects using a joint text embedding space. However, this approach may not be robust for real-world deployment, as the Brain-LLM might not explicitly state the action to perform. To test this, we use a simple BERT embedding matching technique that computes the cosine similarity between high-level plans created by Brain-LLM and the subsequent available list of objects and actions separately. The object and action with the maximum cosine similarity are used to construct the action primitive for a given plan. We demonstrate three examples in Table \ref{tab:bodyllm-examples} where this approach performed incorrectly compared to Body-LLM. Robust planning and successful execution in the real world require accurate low-level controls, which are handled by Body-LLM in our approach. Leveraging powerful LLMs allows for an improved association between natural language plans and low-level control commands.

\section{Experiments with the Franka Robotic Arm}
\label{sec:franka-sim-exp}

\begin{table*}[t]
\caption{BrainBody-LLM Franka Arm Simulation Results. GPT-4 succeeds in all tasks including the difficult ones that require processing more than one possible feedback steps.}
\label{tab:franka-exp}
\centering
\small
\begin{tabular}{@{}lccccc@{}}
\toprule
\multicolumn{1}{c}{\textbf{Task}}                                                                                                                                                                                        & \textbf{\begin{tabular}[c]{@{}c@{}}Unreachable\\ Obj./Loc.\end{tabular}} & \textbf{Diff.} & \textbf{GPT-4} & \textbf{GPT-3.5} & \textbf{PaLM 2} \\ \midrule
\begin{tabular}[c]{@{}l@{}}Move the cubes to the right side of the\\ environment only.\end{tabular}                                                                                                                      & No                                                                       & Easy           & Success        & Success          & Success         \\ \midrule
\begin{tabular}[c]{@{}l@{}}Use the cubes to form the shape of the\\ capital letter ‘L’. The shape consists of\\ a horizontal and a vertical line intersecting\\ at a right angle.\end{tabular}                           & No                                                                       & Medium         & Success        & Success          & Fail            \\ \midrule
\begin{tabular}[c]{@{}l@{}}Arrange cubes in a horizontal line in the\\ center of the work space, ordering them\\ from left to right based on their color in\\ sequence of the visible spectrum.\end{tabular}             & Yes                                                                       & Medium         & Success        & Fail             & Success         \\ \midrule
\begin{tabular}[c]{@{}l@{}}Arrange objects in the scene such that\\ left and right side of the environment has\\ exactly one cube and one cylinder. Each\\ object needs to be assigned a unique\\ location.\end{tabular} & Yes                                                                       & Medium         & Success        & Fail             & Success         \\ \midrule
\begin{tabular}[c]{@{}l@{}}Create a triangle in the right part of the\\ working space using reddish cubes.\end{tabular}                                                                                                  & Yes                                                                       & Hard           & Success        & Fail             & Fail            \\ \midrule
\begin{tabular}[c]{@{}l@{}}Create a plus sign in the working place\\ using the cubes.\end{tabular}                                                                                                                       & Yes                                                                       & Hard           & Success        & Fail             & Fail            \\ \midrule
\begin{tabular}[c]{@{}l@{}}Segregate objects based on geometric\\ shapes into left and right part of work\\ space. Objects of different shapes should\\ not be on the same part of work space.\end{tabular}              & Yes                                                                       & Hard           & Success        & Fail             & Fail            \\ \bottomrule
\end{tabular}
\end{table*}

\begin{figure*}[!h]
\centering
\includegraphics[width=\textwidth]{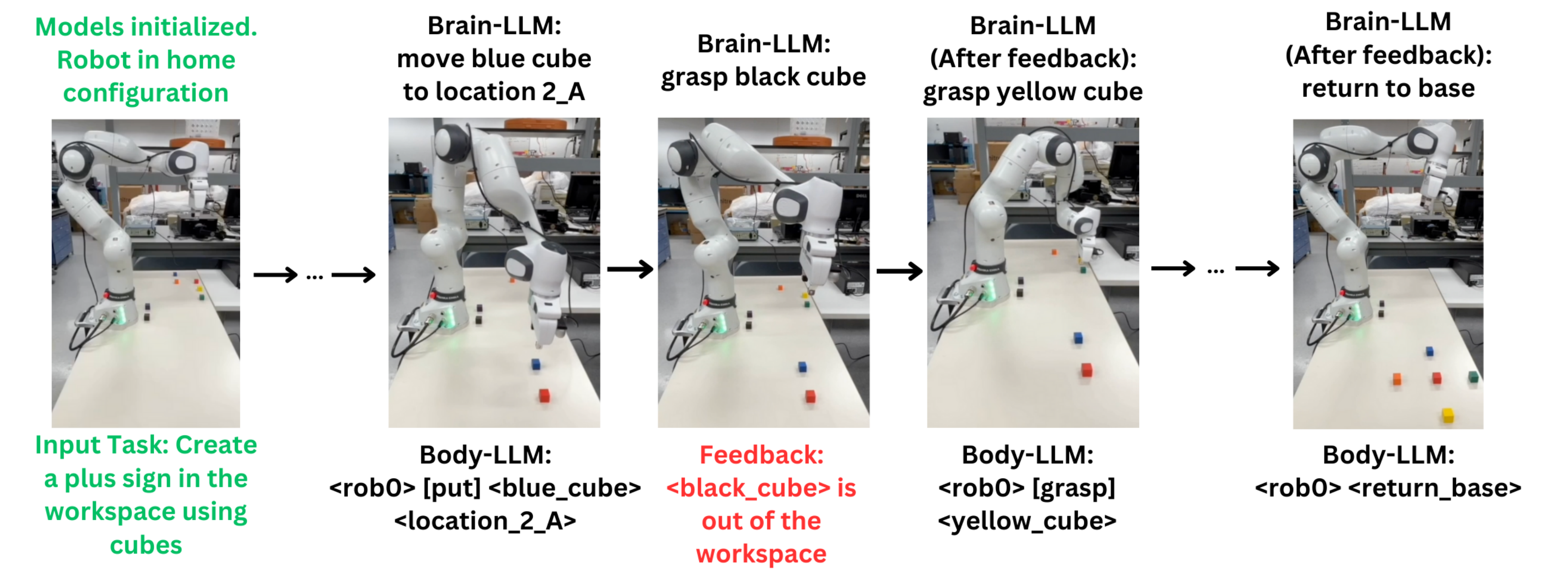}
\caption{Demonstration of our BrainBody-LLM algorithm (GPT-4 - Backend), operating on a Franka Research 3 Robotic Arm. It successfully completes the sixth task in table \ref{tab:franka-exp} by creating a ``Plus" sign right side of the environment and subsequently returning to its initial position.}
\label{fig:real-robot-exp}
\end{figure*}

In this section, we demonstrate how BrainBody-LLM extends to real-world robot task planning using the Franka Research 3 (FR3) robotic arm. Our experiments, integrating a 7 DOF robotic arm with advanced LLMs like PaLM 2, GPT-3.5, and GPT-4, show promising results in autonomous robotic task planning and execution. All our experiments assume that the initial locations of objects and available target locations are known to the LLMs through environmental information communicated via prompts. This assumption can be removed with visual perception modules that identify and locate objects \citep{minderer2022simple}.

\subsection{Task Design}
We designed seven pick-and-place tasks for the Franka arm, utilizing basic shapes such as cubes and cylinders to construct specific configurations in a tabletop setup. During execution, the task description and environmental information about the objects and available target locations in the scene are provided to the Brain-LLM. Some of the objects and target locations in the environment may be unreachable by the arm. The LLM planner must utilize controller errors to correct plans that involve unreachable objects or locations. Therefore, these tasks not only test the precision and spatial reasoning of the task planners but also their adaptability to errors, evaluating the feasibility of an autonomous end-to-end robotic system.

The tasks listed in Table \ref{tab:franka-exp} vary in terms of whether unreachable objects or locations are presented to them. The first two tasks, in which the Brain-LLM is not provided with any unreachable objects or locations, do not necessitate a feedback step. In contrast, it is likely that at least one feedback step is required to complete the remaining tasks. We classify these tasks into three levels of difficulty: Easy, Medium, and Hard, based on the minimum number of feedback iterations needed for successful task completion. Examples used in our prompts are described in Appendix \ref{app:prompts-franka}.

\subsection{Results on Franka Research 3
Simulation Environment}
%Experimental Setup}
We used a popular simulation environment for the FR3, which employs differential optimization and control barrier functions for pick-and-place tasks~\citep{Dai2023SafeNA}. Scene objects were created with MuJoCo~\citep{todorov2012mujoco}, and we developed scripts for API integration with three LLMs—PaLM 2, GPT-3.5, and GPT-4—using the prompt structure described in Section \ref{sec:algo2}. Our tabletop environment consisted of objects distributed in both accessible and inaccessible parts of the workspace. This setup tested the LLM-based planner's ability to adapt to out-of-workspace errors in real-time and revise plans accordingly. Extensive trajectory testing was conducted to avoid singularity points. We used a Python-based control repository\footnote{\url{https://github.com/Rooholla-KhorramBakht/FR3Py}} for the FR3 arm.

Our results show that GPT-4 excelled in all tasks, while GPT-3.5 completed 2 out of 7 tasks, and PaLM 2 achieved 3 out of 7 tasks. All models managed the first, simpler task without feedback loops. Figure \ref{fig:GPT-4-plans-feedback} (RHS) demonstrates one of the tasks successfully completed by our algorithm.

\subsection{Results on the Franka Robotic Arm}
One of the chosen tasks involved arranging cubes to form a plus sign (\textit{Create a plus sign in the working space using the cubes}), testing the planners' logical and spatial reasoning along with their error resolution effectiveness. We selected this task for our real-world experiment since it involved multiple feedback steps and had a higher difficulty level than the other tasks. Figure \ref{fig:real-robot-exp} shows the setup and GPT-4's successful task execution, illustrating its capability to control a 7 DOF robotic arm in object manipulation tasks.

GPT-3.5 struggled with plan adjustments after feedback, often failing to resolve incorrect steps. After utilizing all feedback loops, the incorrect plans were either executed or ignored by the Body-LLM, resulting in a low success rate. PaLM 2 occasionally generated incorrect plans but adeptly revised plans post-feedback, as seen in the third and fourth tasks in Table \ref{tab:franka-exp}. However, not all plans led to an error, and in some tasks, although the execution was completed, the final configuration created did not match the specified task.

While PaLM 2 demonstrated planning and error correction capabilities, and GPT-3.5 showed good spatial reasoning, GPT-4 improved on both these models by effectively utilizing run-time feedback errors for accurate task planning.

\subsection{Guidelines for Adapting BrainBody-LLM to Diverse Robotic Environments}
\label{subsec:recipes-brainbodyllm}

The BrainBody-LLM framework is designed to enable direct autonomous planning across a broad spectrum of robotic tasks, both in simulation and real-world environments. Apart from our experiments in VirtualHome, datasets such as ALFRED~\citep{ALFRED20}, VRKitchen~\citep{VRKitchen}, and TEACh~\citep{Padmakumar2021TEAChTE} also employ high-level plans for task execution, making them ideal candidates for integration with our framework. Given that prompt-based experiments can be time-consuming, we propose a streamlined yet effective approach for adapting BrainBody-LLM to custom robotic tasks and environments:

\begin{enumerate}
    \item \textbf{Data and prompt preparation}:
    \begin{enumerate}
        \item \textit{Prepare in-context learning examples.} Develop examples for the prompt frameworks defined in Section \ref{sec:algo2}. The planning phase (BrainLLM) includes environment setup, task name, and high-level plans. The execution phase (BodyLLM) involves translating high-level plans into low-level controls. The feedback phase (BrainLLM) encompasses error messages, explanations, and updated plans. Examples of data tuples for each phase are provided in Appendix \ref{app:prompts-vh} and Appendix \ref{app:prompts-franka} for VirtualHome and Franka experiments respectively.
        
        \item \textit{Incorporate prepared examples in prompting frameworks.} Add collected examples to the prompts of structures defined in Figures \ref{fig:planning-prompt}, \ref{fig:execution-prompt}, and \ref{fig:feedback-prompt} to create final prompts for the LLM API call.
    \end{enumerate}
    
    \item \textbf{Task specific modifications}:
    \begin{enumerate}
        \item \textit{Define the environment.} Specify the list of objects and actions available to the robot. In our prompts, the variables \textit{object\_list} and \textit{actions\_available} are used to ground the LLM in the environment, minimizing hallucinations.
        
        \item \textit{Incorporate special instructions.} Depending on the task requirements, it may be necessary to include specific instructions to ensure that the LLM adheres to the environment's constraints. PaLM 2 and GPT-3.5 are particularly sensitive to prompt variations. For example, emphasizing critical instructions using exclamation points and bold formatting enhances the LLM's performance.
    \end{enumerate}
    
    \item \textbf{General guidelines for prompt engineering in robotic task planning}:
    \begin{enumerate}
        \item \textit{Optimize prompt length.} Longer prompts can cause the LLM to lose important context and increase the cost of API calls. These issues are largely mitigated in GPT-4, which offers advanced context retention and greater parameter complexity.
        
        \item \textit{Customize for real-world tasks.} Adapting BrainBody-LLM for real robotic tasks necessitates human-curated examples and customization tailored to the specific task at hand.
    \end{enumerate}
\end{enumerate}

\section{Conclusion}
\label{sec:conclusion}

This paper introduced an algorithm for robotic task planning that leverages the reasoning and dynamic error correction capabilities intrinsic to LLMs. Our approach improves the success rate of task execution and recall of goal conditions, positioning it favorably against current baselines within the Virtual Home Human-In-A-Household simulation environment. The design and conceptual framework of our algorithm are practical and readily adaptable for real-world implementation, as demonstrated through our experiments using the Franka Research 3 Robotic Arm in Section \ref{sec:franka-sim-exp}.

The assumption in our work regarding the reversibility of errors made by the robot introduces a notable limitation. The framework relies on the ability to create a new plan for achieving the given task after receiving environmental information and an error message. However, errors made by the planner might irreversibly alter the environment or damage the robot hardware, making the task unachievable. In future work, we aim to address this limitation by incorporating simulators into the planning loop and designing safety guarantees for the underlying action primitives used by Body-LLM. Our approach relies on implicit intervention through simulators or other forms of feedback communication with the environment. In the future, we aim to evaluate our approach in more complex and long-horizon scenarios compared to the experiments presented in this paper.

Future work will focus on understanding and mitigating the phenomena of closed-loop oscillations and hallucinations observed in LLM-generated plans. This effort will involve leveraging multi-modal feedback through diverse sensor modalities to ensure grounded and realistic task planning. Our overarching goal is to evaluate and refine the efficacy of LLM-based planning algorithms, thereby facilitating their broader adoption in robotics applications.

\noindent \textbf{Acknowledgements} This work was supported in part by ARO under Grant W911NF-22-1-0028 and in part by the New York University Abu Dhabi (NYUAD) Center for Artificial Intelligence and Robotics (CAIR), funded by Tamkeen under the NYUAD Research Institute Award CG010.

\noindent \textbf{Author Contributions} Conceptualization, V.B., A.K., P.K. and F.K.; methodology, V.B., A.K. and P.K.; software, V.B. and A.K.; resources, P.K., R.K., and F.K.; writing---original draft preparation, V.B. and A.K.; writing---review and editing, V.B., A.K., P.K., R.K. and F.K.; project administration, P.K., R.K. and F.K.; funding acquisition, P.K., R.K. and F.K.

\noindent \textbf{Data availability} No datasets were generated during the current study.

\noindent \textbf{Competing interests} The authors declare no competing interests.

\noindent \textbf{Ethics Approval} All authors consented to the submission of this manuscript and we explicitly state that the presented work is original and not submitted anywhere else. Our submission follows all ethical guidelines as outlined in the Autonomous Robots journal.

\bibliography{sn-bibliography}% common bib file


\appendix
\section{Prompt Examples for VirtualHome Experiments}
\label{app:prompts-vh}

\begin{figure}[H]
  \centering
  \small
  \begin{tcolorbox}[title=Planning prompt of Brain-LLM for VirtualHome Experiments]
You are in the command of a virtual agent...

Some examples of High-level Instruction - Step-by-step Plans pairs are given below.\\

\textbf{High-level Instruction:} refrigerate the salmon\\
\textbf{Step-by-step Plans:}\\
0: I would go to the kitchen and find the salmon.\\
1: I would take the salmon and put it in the fridge.\\
2: I would close the fridge.\\
3: Done.\\

\textbf{High-level Instruction:} turn off the table lamp\\
\textbf{Step-by-step Plans:}\\
0: walk to the table lamp.\\
1: find the switch.\\
2: turn off the switch.\\
3: Done.\\

\textit{ ... (More examples) ...}\\
\\
You have the following objects in scene: \{object\_list\}. The list of available actions are - walk, run, walktowards, walkforward, turnleft, turnright, sit, standup, grab, open, close, put, putin, switchon, switchoff, drink, touch and lookat.\\ 
$\cdots$
            \end{tcolorbox}
\end{figure}

\begin{figure}[H]
  \centering
  \small
  \begin{tcolorbox}[title=Feedback prompt of Brain-LLM for VirtualHome Experiments]
  You are in the command of a virtual agent...
  
    Some examples of high level tasks, generated subtasks, error step and error message are given below...
    
    \textbf{Initial Plan:}\\
    0: Walk to the radio.\\
    1: Look at the radio. \\
    2: Grab the radio. \\
    3: Switch off the radio. \\
    4. Done. \\
    \textbf{Error Step:} 2 \\
    \textbf{Feedback:} {{`0': {{`message': `ScriptExcutor 0: execution\_general: Script is impossible to execute'}}}}\\
    Explanation: Radio is an object that cannot be grabbed. You can either turn it off or turn it on.\\ 
    \textbf{New Plan:}\\
    2: Switch off the radio.\\
    3. Done.\\
    
    \textit{ ... (More examples) ...}\\
    
    You were given task: \{input\}, a generated Initial Plan was: \{init\_plan\}. The robot received the following feedback message: \{feedback\_message\} \\

    You have the following objects in scene: {object list}. The list of available actions are - walk, run, walktowards, walkforward, turnleft, turnright, sit, standup, grab, open, close, put, putin, switchon, switchoff, drink, touch and lookat. \\
    $\cdots$
  \end{tcolorbox}
\end{figure}

\begin{figure}[H]
  \centering
  \small
  \begin{tcolorbox}[title=Execution prompt of Body-LLM for VirtualHome Experiments]

  You are in control of a virtual agent...
  
Some examples of plan - action program pairs are given below - 

\textbf{Description:} walk to laundry room\\
\textbf{Action Plan:} \textless char0\textgreater  [walk] \textless laundryroom\textgreater \\
\textbf{Explanation:} Since there exists a walk action that is executable by the simulator and a bathroom in the simulator, this action plan will satisfy the given description.\\

\textbf{Description:} put clothes in washing machine\\
\textbf{Action Plan:} \textless char0\textgreater  [putin] \textless clothespile\textgreater  \textless washingmachine\textgreater \\
\textbf{Explanation:} Since there exists a putin action that is executable by the simulator, and a washing machine and clothespile in the simulator, this action plan will satisfy the given description.\\

\textbf{Description:} add detergent \\
\textbf{Action Plan:} \textless pass\textgreater\\
\textbf{Explanation:} There exists detergent in the scene but there is no action add that is executable by the simulator so I should not generate an action plan but simply pass it.\\

\textit{ ... (More examples) ...}\\

You have the following objects in scene: \{object\_list\}. The list of available actions are: walk, run, walktowards, walkforward, turnleft, turnright, sit, standup, grab, open, close, put, putin, switchon, switchoff, drink, touch, lookat

Return a suitable action program for the provided plan...

\textbf{Description:}  \{input\} \\
\textbf{Action Plan:}
  \end{tcolorbox}
\end{figure}

\newpage
\section{Prompt Examples for Franka Arm Experiments}
\label{app:prompts-franka}

\begin{figure}[!h]
  \centering
  \small
  \begin{tcolorbox}[title=Planning Prompt of Brain-LLM for Franka Arm Experiments]

  You are in the command of the Franka Arm...

Some examples of High-level Instruction - Step-by-step Plan pairs are given below --

    \textbf{High-level Instruction:} separate the blue cubes from the others.\\
    \textbf{Step-by-step Plans:}\\
      0: grasp blue cube \\
      1: move blue cube to location\_A \\
      2: grasp aqua cube \\
      3: move aqua cube to location\_B \\
      4: grasp azure cube \\
      5: move azure cube to location\_C \\
      6: grasp yellow cube \\
      7: move yellow cube to location\_D \\
      8: grasp beige cube \\
      9: move beige cube to location\_E\\
      10: return to base \\

    \textbf{High-level Instruction:} order the rainbow colors on the table \\
    \textbf{Step-by-step Instructions:}\\
    0: grasp red cube\\
    1: move red cube to location\_A\\
    2: grasp orange cube\\
    3: move orange cube to location\_B\\
    4: grasp yellow cube\\
    5: move yellow cube to location\_C\\
    6: grasp green cube\\
    7: move green cube to location\_D\\
    8: grasp blue cube\\
    9: move blue cube to location\_E\\
    10: grasp indigo cube\\
    11: move indigo cube to location\_F\\
    12: grasp violet cube\\
    13: move violet cube to location\_G\\
    14: return to base \\

    \textit{ ... (More examples) ...}\\

    You have the following objects in scene: \{object\_list\}. The list of available actions are - grasp, move, return to base.\\ 
$\cdots$
  \end{tcolorbox}
\end{figure}

\begin{figure*}[!h]
  \centering
  \begin{tcolorbox}[title=Feedback prompt of Brain-LLM for Franka Experiments]
    You are in the command of a virtual agent...
  
    Some examples of high level tasks, generated subtasks, error step and error message are given below...
    
    \textbf{High-level Instruction:} move 2 of the blueish cubes to the right of the table.\\
    \textbf{Initial plan:}
    0: grasp blue cube \\
    1: move blue cube to location\_A \\
    2: grasp aqua cube \\
    3: move aqua cube to location\_B \\
    4: return to base \\
    \textbf{Error step:} 2 \\
    \textbf{Feedback:} aqua\_cube is out of the workspace of the robot. \\
    \textbf{New plan:} \\
      2: grasp azure cube \\
      3: move azure cube to location\_B \\
      4: return to base \\

    \textbf{High-level Instruction:} order the rainbow colors in the table.\\
    \textbf{Initial plan:}\\
    0: grasp red cube \\
    1: move red cube to location\_A \\
    2: grasp orange cube \\
    3: move orange cube to location\_B \\
    4: grasp yellow cube \\
    5: move yellow cube to location\_C \\
    6: grasp yellow cube \\
    7: move green cube to location\_D \\
    8: grasp blue cube \\
    9: move blue cube to location\_E \\
    10: grasp indigo cube \\
    11: move indigo cube to location\_F \\
    12: grasp violet cube \\
    13: move violet cube to location\_G \\
    14: return to base \\
    \textbf{Error step:} 13 \\
    \textbf{Feedback:} location\_G is out of the workspace of the robot. \\
    \textbf{New plan:} \\
      13: move violet cube to location\_H \\
      14: return to base \\

    \textit{ ... (More examples) ...}\\

    You were given task: \{input\}, a generated Initial Plan was: \{init\_plan\}. The robot received the following feedback message: \{feedback\_message\} \\

    You have the following objects in scene: {object list}. The list of available actions are - grasp, move, return to base. \\
    $\cdots$
  \end{tcolorbox}
\end{figure*}

\begin{figure}[!h]
  \centering
  \small
  \begin{tcolorbox}[title=Body-LLM prompt for Franka Arm Experiments]
        You are in control of a virtual agent...
  
Some examples of plan - action program pairs are given below -

    \textbf{Description:} grasp the white cube \\
    \textbf{Action Plan:} \textless rob0\textgreater [grasp] \textless white\_cube\textgreater\\
    \textbf{Explanation:} Since there exist a grasp action in the repertoire of the controller and white\_cube in the environment, this action plan will satisfy the given description.\\

    \textbf{Description:} hold the black tube \\
 \textbf{Action Plan:} \textless rob0\textgreater [grasp] \textless black\_tube\textgreater\\
    \textbf{Explanation:} Since there exist a grasp action in the repertoire of the controller and black\_tube in the environment, this action plan will satisfy the given description.\\

\textit{ ... (More in-context examples) ...}\\

You have the following objects in scene: \{object\_list\}. The list of available actions are: grasp, move, return to base.

Return a suitable action program for the provided plan...

\textbf{Description:}  \{input\} \\
\textbf{Action Plan:}
  \end{tcolorbox}
\end{figure}

\end{document}